\pdfoutput=1

\documentclass[11pt]{article}
\usepackage{CJKutf8}

\usepackage[final]{acl}
\usepackage{acl}

\usepackage{times}
\usepackage{latexsym}
\usepackage{makecell}
\usepackage[T1]{fontenc}

\usepackage[utf8]{inputenc}

\usepackage{microtype}

\usepackage{algorithm}
\usepackage{algorithmic}
\usepackage{booktabs}
\usepackage{graphicx}
\usepackage{amssymb}
\usepackage{bbold}
\usepackage{bbm}
\usepackage{amsmath}
\usepackage{url}
\usepackage[switch]{lineno}  %
\usepackage{newfloat}
\usepackage{float}
\usepackage{subfig}
\usepackage{multicol}
\usepackage{multirow}
\usepackage{pifont}

\title{Accelerating Dense LLMs via L0-regularized Mixture-of-Experts}

\author{
Zhenyu Zhang \textsuperscript{1}, 
Jiudong Yang \textsuperscript{2}, 
Zhaowen Tao \textsuperscript{1}, 
Meng Chen \textsuperscript{3}\thanks{~ Corresponding author.} \\
\textsuperscript{1} YZW, Chengdu, China\\
\textsuperscript{2} FuTu AI, Shenzhen, China\\
\textsuperscript{3} Wise AI, Melbourne, Australia\\
\tt{zhangzhenyu13@outlook.com, simonyang@futunn.com} \\
\tt{taozhaowen@gmail.com, chenmengdx@gmail.com}
}
\begin{document}
\maketitle

\begin{abstract}

Large language models (LLMs) achieve strong performance but suffer from slow and costly inference. Existing acceleration methods often lead to noticeable performance degradation, while Mixture-of-Experts (MoE) models require extensive computational resources. In this paper, we propose L0-MoE, a lightweight MoE approach using L0-regularization to accelerate dense LLMs nearly without performance loss. Our method introduces a cluster confusion matrix for domain-aware dataset curation and applies dynamic batching for efficient training. Experiments show that L0-MoE achieves up to 2.5x speedup over dense models while maintaining competitive performance, outperforming existing LLM acceleration baselines. Code is now released \footnote{https://github.com/zhangzhenyu13/L0-MOE}.
\end{abstract}

\section{Introduction}


Large language models (LLMs) have demonstrated remarkable intelligence across various tasks \cite{openai2024gpt4technicalreport, geminiteam2024geminifamilyhighlycapable, llam3-dubey2024llama, mistral-jiang2023mistral, deepseekai2025deepseekr1incentivizingreasoningcapability, qwen2}, including question answering, mathematics, coding, and content generation. A key insight into their success is the parameter scaling law \cite{kaplan2020scalinglawsneurallanguage}, which suggests that increasing model size enhances performance across diverse tasks, potentially advancing artificial general intelligence (AGI) \cite{bubeck2023sparksartificialgeneralintelligence}. However, larger LLMs incur high inference costs, leading to slower generation speeds and increased computational expenses. Thus, optimizing LLM inference efficiency has become a critical challenge for both academia and industry.


Various approaches have been proposed to accelerate LLM inference, which can be categorized into three main techniques: (1) \textbf{Quantization}, including GPTQ \cite{frantar2022gptq}, SmoothQuant \cite{smoothquant-xiao2023smoothquant}, AWQ \cite{awq-lin2024awq} and DuQuant \cite{lin2024duquant}, reduces precision by converting weights and activations from floating-point to lower-bit integer formats, significantly improving efficiency. (2) \textbf{Model pruning}, such as LLM-Pruner \cite{llmpruner-ma2023llm} and LLM-Shearing \cite{llmshearing-xia2023sheared}, removes redundant parameters based on predefined criteria to compress models and accelerate inference. (3) \textbf{Knowledge distillation} \cite{rkl-gu2024minillm, cotdistill-feng2024teaching}, like reverse-KD \cite{rkl-gu2024minillm} and Chain-of-Thought (CoT) Distillation \cite{cotdistill-feng2024teaching}, transfers knowledge from large LLMs to smaller ones using distillation techniques \cite{hinton2015distillingknowledgeneuralnetwork}, reducing computational demands. While these methods achieve substantial speedup, they often come at the cost of performance degradation, posing challenges for real-world deployment.

\begin{figure*}[ht]
    \centering
    \setlength{\abovecaptionskip}{1mm}
    \setlength{\belowcaptionskip}{-6mm}
    \includegraphics[width=0.85\textwidth]{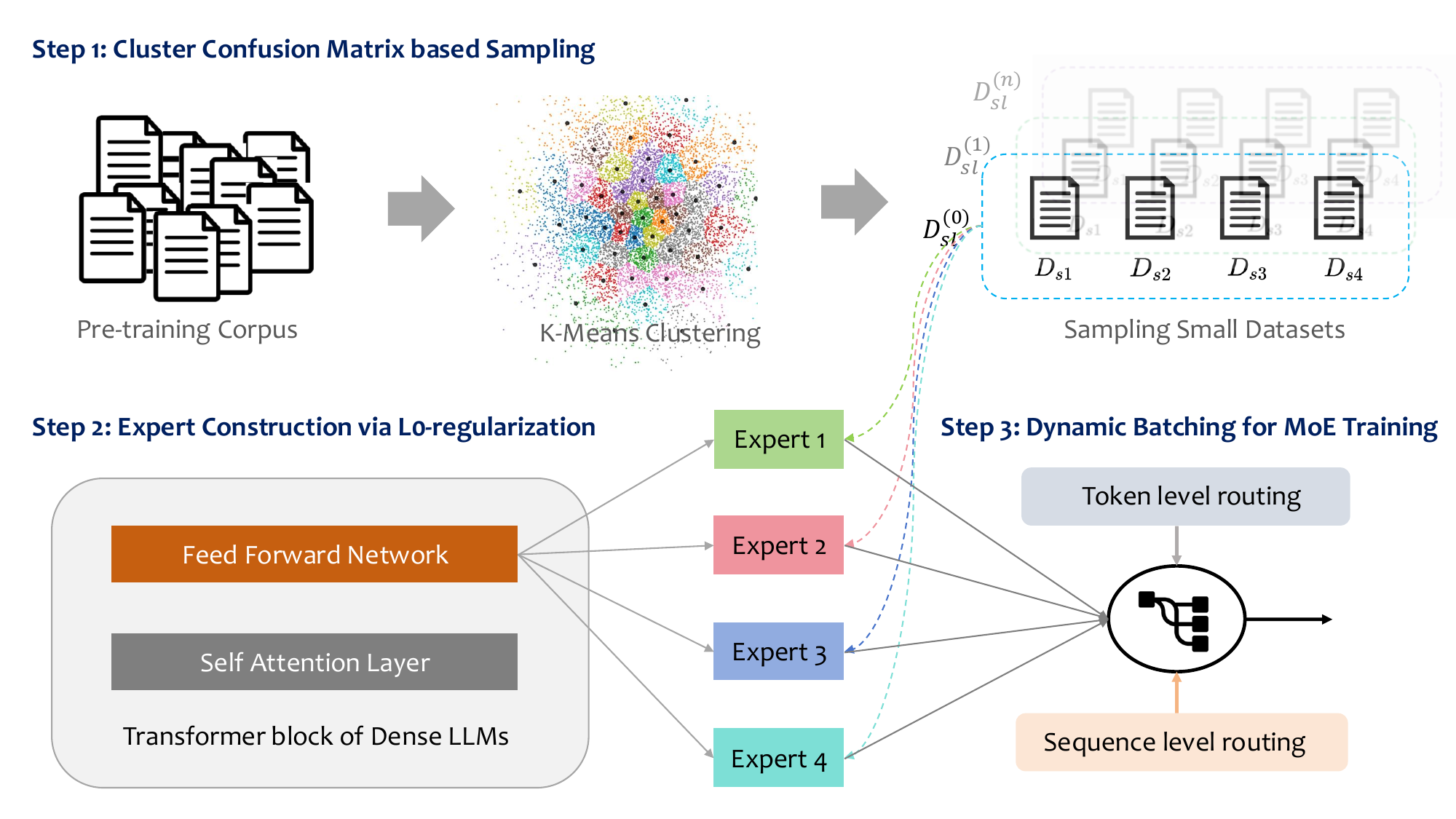}
    \caption{Overview of the L0-MoE Architecture, which includes three main stages: (1) cluster confusion matrix based sampling, (2) expert formation using L0 regularization, and (3) dynamic batching for MoE training. The figure above illustrates the process of building an L0-MoE with four experts over $n$ iterations of dataset sampling.}
    \label{fig:guidance}
\end{figure*}

Recently, sparsely gated Mixture-of-Experts (MoE) models \cite{moesurvey-cai2024survey}, particularly in transformer-based large language models, have significantly improved inference speed optimization. MoE operates on a simple yet effective principle: different model components, known as experts, specialize in distinct tasks or data aspects. For a given input, only relevant experts are activated, reducing computational costs while leveraging a vast pool of specialized knowledge. This scalable and flexible approach aligns with the scaling law, enabling larger model capacities without proportional computational overhead. However, current MoE training focuses on training from scratch or upcycling dense LLMs, both requiring vast computational resources and high-quality corpora. For instance, DeepSeek-V3 \cite{deepseekai2024deepseekv3technicalreport} and Qwen2.5-Max \cite{qwen2025qwen25technicalreport} were pre-trained on 14.8T and 20T tokens, respectively, with additional fine-tuning, making them costly and less accessible. In contrast, \textit{little research has explored leveraging MoE to accelerate inference using a small-scale training corpus (e.g., tens of billions of tokens) while maintaining performance comparable to dense LLMs}. This direction is particularly appealing for large-scale industrial applications with cost-sensitive deployment constraints.

To address this issue, we propose \textbf{L0-MoE}, a mixture-of-experts (MoE) model built via L0-regularization \cite{l0reg-louizos2017learning} using a small, curated \textbf{30B}-token corpus. Our approach has two key components: (1) L0-regularization selects critical hidden dimensions in transformer MLPs to form experts. (2) A cluster confusion matrix (CCM)-based sampling method curates the training corpus and schedules dynamic batching. Using the BGE-M3 encoder \cite{bgem3-chen2024bge} and K-means clustering \cite{Jin2010}, we extract diverse semantic domains from RedPajama \cite{weber2024redpajama} to construct expert-relevant sub-datasets. A gating mechanism and dynamic batching optimize training. L0-MoE achieves \textbf{2.5×} inference speedup with no obvious performance loss across four benchmarks. Our contributions are as follows: 1) We introduce a novel MoE building method leveraging L0-regularization, enabling efficient LLM inference acceleration with minimal training cost. 2) We propose a CCM-based corpus curation and dynamic batching strategy for effective MoE training. 3) Extensive experiments validate the efficiency of our method in achieving inference speedup while maintaining performance.

\vspace{-5pt}
\section{Preliminary}
\vspace{-5pt}
\subsection{L0-regularization}

L0-regularization \cite{l0reg-louizos2017learning} is a powerful technique for feature selection and parameter pruning in neural networks. It imposes a penalty on parameters that deviate from zero, without additional constraints. This approach enhances model efficiency by eliminating unnecessary computations and resources, as irrelevant parameters are pruned and thus not computed. For a given weight matrix $W \in R^{m \times n}$, a mask matrix $Z \in {0,1}^n$ is employed to derive a reduced weight $g(W,Z) \in R^{m \times n0}$, where $g$ selects $n0 < n$ columns from $W$ using $Z$. Due to the non-differentiable nature of $Z$, optimizing it is challenging. To address this, the binary hard concrete function is introduced for L0-regularization, as shown in Equation \ref{eq:l0-reg}.
\setlength{\abovedisplayskip}{3pt}
\setlength{\belowdisplayskip}{3pt}
\begin{equation}
\label{eq:l0-reg}
    \begin{matrix}
u \sim \mathcal{U} 
 \\
s= Sigmoid( (log(u) - log(1-u) + log$a$ ) / $b$ )
 \\
\bar{s} = s(\zeta - \gamma ) + \gamma 
 \\
z= min(1, max(0, \bar{s} ))
\end{matrix}
\end{equation}


The uniform distribution $\mathcal{U}$ is defined over the interval [0,1]. We set the hyper-parameters as $b=0.83$, $\zeta=1.1$, and $\gamma=-0.1$ by following \citet{l0reg-louizos2017learning}. Using the learned $z$, we estimate the proportion of retained weights as $\hat{r}=\frac{sum(z)}{m*n}$. To effectively control the desired retention ratio $r$ for a given weight matrix $W$, we employ a Lagrangian multiplier \cite{l0lagmul-wang2019structured}, as described in Equation \ref{eq:lag-mul}.
\setlength{\abovedisplayskip}{4pt}
\setlength{\belowdisplayskip}{3pt}
\begin{equation}
\label{eq:lag-mul}
    \mathcal{L}_{l_0} = \lambda_1 (\hat{r} -r ) + \lambda_2(\hat{r} - r )^2
\end{equation}


We initialize the learnable parameters $\lambda_1$ and $\lambda_2$ to 0 in our experiments. In our approach, $r$ represents the retention ratio of the feed-forward network (FFN) up-projection dimension.

\subsection{Mixture of Expert}
\vspace{-2pt}
Mixture of Experts (MoE) \cite{moesurvey-cai2024survey} employs a modular architecture comprising a gating network and multiple expert networks to enhance efficiency and performance through parameter scaling. This architecture partitions the model into several experts, each specializing in specific subsets of input data. MoE utilizes a gating mechanism with a router to dynamically select the appropriate experts for processing incoming inputs, allowing the model to concentrate on relevant features while minimizing unnecessary computations. In our approach, the router is implemented as a linear projection layer $W_{router} \in R^{d \times N}$. MoE incorporates two auxiliary losses (Equation \ref{eq:aux-loss}), such as the load balancing loss $\mathcal{L}_{balance}$ \cite{loadblmoe:fedus2022switch} and the router $Z$-Loss $\mathcal{L}{z}$ \cite{zlossmoe-zoph2022st}, to promote a balanced distribution of inputs among experts. These losses penalize high values in the logits produced by the gating network, encouraging a more even allocation of tokens to experts.
\begin{equation}
\label{eq:aux-loss}
    \begin{matrix}
\mathcal{L}_{aux}= \mathcal{L}_{balance} + \lambda\mathcal{L}_{z}
\\
\mathcal{L}_{balance} = \sum_{i=1}^{i=N} (\frac{c_i}{B} - \frac{1}{N}  )^2
\\
\mathcal{L}_{z} =\frac{1}{B} \sum_{1}^{B}(log(\sum_i^{N}e^{x_i^{(j)}} )  )
\end{matrix}
\end{equation}

Here $c_i$ represents the tokens of the $i^{th}$ expert, and $N$ denotes the number of experts. The batch contains $B$ tokens. The logit for the $j^{th}$ token from the $i^{th}$ expert, as determined by the router module, is denoted as $x_i^{(j)}$.

\vspace{-5pt}
\section{Approach}
\vspace{-5pt}
\subsection{Cluster Confusion Matrix based Sampling}
\label{sec:ccm}

Given a pretraining corpus, we construct training datasets via the following steps: 1) Randomly sample a small subset without replacement and use the BGE-M3 encoder \cite{bgem3-chen2024bge} to extract $d_{sv}$-dimensional semantic vectors for each sample. 2) Apply the K-means clustering algorithm \cite{Jin2010} to the semantic vectors to identify $K$ centers $C \in \mathbb{R}^{K \times d_{sv}}$. Divide the small subset into $K$ folds and sample $m$ instances from each fold to form a dataset $D_{sl}=\{D_{s_1}, \ldots, D_{s_K}\}$ for domain semantic learning, where $|D_{s_k}|=m$ for $1 \le k \le K$. 3) Repeat steps 1 and 2 for $Q$ iterations to obtain $Q \times K$ centers and $Q$ datasets. For the $l^{th}$ iteration ($l=\{1,2,\ldots,Q\}$), the cluster centers are $C^{(l)} \in \mathbb{R}^{K \times d_{sv}}$ and the constructed dataset is $D_{sl}^{(l)}$.


We define the clustering confusion matrix (CCM) as per Equation \ref{eq:ccm}, where $\delta=0.1$ is a hyperparameter, $C_{i}$ represents the $i^{th}$ center vector, and $CCM[i,l]$ denotes the clustering confusion value for the $i^{th}$ center at iteration $l$. The hypothesis posits that the semantic domain distance for the $i^{th}$ center between $C_i$ and $C_i^{(l)}$ can be assessed using bidirectional inter-clustering ($f_1$ and $f_2$) and intra-clustering ($f_3$) cosine similarity. A larger semantic domain distance indicates that $D_{sl}^{(l)}$ from the $l^{th}$ iteration divides domains more distinctly. We compute the domain semantic distance using Equation \ref{eq:dsd} and reorder the $Q$ datasets based on $d_{ds}^{(l)}$. In addition to the initial $D_{sl}^{(0)}$, our datasets now include $Q-1$ ordered datasets $D_{sl}^{ord(l)}$, where $ord(l)$ is the order index.
\setlength{\abovedisplayskip}{4pt}
\setlength{\belowdisplayskip}{4pt}
\begin{equation}
\label{eq:ccm}
    \begin{matrix}
CCM[i,l] = (f_1 + f_2) * f_3
\\
f_1(i,l) =\frac{1}{K}  \sum_{k=1}^{k=K} e^{1- sim(C_i, C_k^{(l)})}
\\
f_2(i,l) =\frac{1}{K}  \sum_{k=1}^{k=K} e^{1-sim(C_k, C_i^{(l)})}
\\
f_3(i,l) = \delta \frac{\sum_{k=1}^{k=K} e^{sim(C_k^{(l)}, C_i^{(l)})} }{\sum_{k=1}^{k=K} e^{sim(C_k, C_i)}}  
\end{matrix}
\end{equation}
\setlength{\abovedisplayskip}{3pt}
\setlength{\belowdisplayskip}{3pt}
\begin{equation}
\label{eq:dsd}
    d_{ds}^{(l)}=max(CCM[:,l]) + \frac{\beta  }{K} \sum_{i=1}^{i=K}CCM[i,l]
\end{equation}
\subsection{Expert Construction via L0-regularization}
\label{sec:expert-construct}
We construct the experts using pretrained checkpoints of dense LLMs. The intermediate size of the feed forward network (FFN) layer is $d_{int}$, and we apply a mask $Z \in R^{d_{int}}$. For each domain subset $D_{s_k}$ ($k \in \{1, 2, \ldots, K\}$) derived from the initial $D_{sl}^{(0)}$, we employ the LLM pretraining loss $\mathcal{L}_{llm}$ along with the L0-regularization loss, as specified in Equation \ref{eq:l0-reg}, to select $r \in (0,1)$*100\% of the dimensions from $d_{int}$, following Equation \ref{eq:train-expert}.
\setlength{\abovedisplayskip}{3pt}
\setlength{\belowdisplayskip}{3pt}
\begin{equation}
    \label{eq:train-expert}
    \mathcal{L}_{exp} = \mathcal{L}_{llm} + \mathcal{L}_{l0}
\end{equation}
To ensure stable training, we gradually adjust $r$ from 100\% to the target ratio $r^{target}$. We freeze all non-MLP parameters of dense LLMs, and the L0-regularization-based training yields $K$ experts, each specialized for distinct semantic domains.

\subsection{Dynamic Batching for MoE Training}
\label{sec:dynamic-batching}
To train the MoE to effectively select appropriate experts based on inputs, we follow Equation \ref{eq:train-moe}, where $\mathcal{L}_{aux}$ is defined in Equation \ref{eq:aux-loss}. The MoE is initialized with $K$ pre-trained experts and a router for each MoE layer.
\setlength{\abovedisplayskip}{3pt}
\setlength{\belowdisplayskip}{3pt}
\begin{equation}
    \label{eq:train-moe}
    \mathcal{L} = \mathcal{L}_{llm} + \alpha\mathcal{L}_{aux}
\end{equation}


We employ a two-loop batch construction strategy during training: 1) domain semantic distance scheduling, where we begin with $D_{sl}^{ord(l)}$ having a lower $d_{ds}$; 2) multi-domain gathering scheduling, where samples in $D_{sl}^{ord(l)}$ are arranged in a cyclic sequence order $x_i^{1}, x_i^{2}, \ldots, x_i^{K}$, and we select $p*K$ ($p=\{1,2,3,\ldots\}$) samples to form a batch. This scheduling offers two advantages: 1) In the initial iterations, the MoE rapidly learns to select appropriate experts since the domain samples in $D_{sl}$ have been previously encountered by the experts. Consequently, the sequence-level selection capabilities of routers are effectively initialized. 2) As training progresses, the domains in $D_{sl}^{ord(l)}$ gradually transition to different semantic spaces, encouraging routers to select multiple experts for each input sample. This enhances the token-level selection capabilities of the routers.

\vspace{-4pt}
\section{Experiments}
\vspace{-5pt}
\subsection{Experimental Setup}

\begin{table*}[htp]
\centering
\small
\begin{tabular}{c|cccccc}
\hline 
{Model} & {MMLU} & {GSM8K} & {HumanEval} & {BBH} & {Average} & {Speedup} \cr
\hline
{Llama-3-8B } & {\textbf{66.6}} & {\textbf{56.0}} & {33.5} & {\textbf{57.7}} & {\textbf{53.5}} \cr
{Llama-3-8B w/ L0-MoE } & {66.3} & {55.9} & {\textbf{33.7}} & {57.2} & {53.3} & {2.0x} \cr
\hline
{Mistral-7B } & {64.1} & {52.2} & {29.3} & {\textbf{56.1}} & {50.4} \cr
{Mistral-7B w/ L0-MoE} & {\textbf{64.8}} & {\textbf{53.6}} & {\textbf{31.1}} & {55.9} & {\textbf{51.4}} & {2.1x} \cr
\hline
{Qwen2-7B } & {70.3} & {79.9} & {51.2} & {\textbf{62.6}} & {66.0} \cr
{Qwen2-7B w/ L0-MoE} & {\textbf{70.4}} & {\textbf{80.5}} & {\textbf{52.0}} & {61.5} & {\textbf{66.1}} & {\textbf{2.5x}} \cr
\hline
\end{tabular}
\caption{Evaluation of different LLMs on MMLU, GSM8K, HumanEval and BBH benchmarks.}
\label{tab:exp-base}
\vspace{-5mm}
\end{table*}

\noindent \textbf{Dataset.} We train on the RedPajama dataset \cite{weber2024redpajama}, a replicated pre-training corpus for LLaMA models, following prior work \cite{llmshearing-xia2023sheared}. Evaluation is conducted on four public benchmarks: MMLU \cite{hendrycks2021measuring}, GSM8K \cite{cobbe2021trainingverifierssolvemath}, HumanEval \cite{chen2021evaluatinglargelanguagemodels}, and BigBench Hard (BBH) \cite{suzgun-etal-2023-challenging}. Each benchmark evaluates distinct aspects of model performance, offering insights into the strengths and limitations of LLMs. 

\noindent \textbf{Baselines.} To assess effectiveness and versatility, we evaluate our method on three open-source LLMs: Llama-3-8B \cite{llam3-dubey2024llama}, Mistral-7B \cite{mistral-jiang2023mistral}, and Qwen2-7B \cite{qwen2}. Comparisons include L0-regularized MoEs, original LLMs, and inference optimization techniques such as GPTQ quantization \cite{frantar2022gptq}, LLM Shearing pruning \cite{llmshearing-xia2023sheared}, and RKD + CoT knowledge distillation \cite{rkl-gu2024minillm, cotdistill-feng2024teaching}.  For CCM, we run 21 iterations, collecting 30B tokens. Experiments use a cluster/expert size of $K=64$ with linear warmup, annealing, and a peak learning rate of 1e-4. Further details are in Appendix \ref{appendix:experimental-setups}.

\noindent \textbf{Implementation Details.} We train our model using the FSDP framework\footnote{\url{https://pytorch.org/docs/stable/fsdp.html}}, employing a layer-wise wrapping policy with the Zero-3 parameter sharding strategy, without CPU offloading. For inference during evaluation, we utilize the SGlang framework\footnote{\url{https://github.com/sgl-project/sglang}}, which is highly optimized for the efficient execution of both dense LLMs and MoEs. All baseline models in our experiments utilize the same SGlang inference framework, ensuring a fair and consistent comparison of inference speeds. Our method is framework-agnostic and can similarly be implemented using other inference frameworks (e.g., vLLM\footnote{\url{https://github.com/vllm-project/vllm}}). The primary source of inference acceleration in our work is the proposed L0-regularization-based MoE architecture, not the inference framework itself. To ensure a fair comparison, we strictly adhere to the original evaluation settings for each benchmark. To support future research, we will release our curated dataset and code to enhance the reproducibility of our work.

\subsection{Main Results}


Table \ref{tab:exp-base} presents the model with the highest performance under our settings. The L0-MoE consistently achieves a $2\text{-}2.5\text{x}$ inference speedup across all base LLMs. Additionally, L0-MoE maintains performance comparable to the base LLMs across four benchmarks, with the L0-MoE variant of Mistral even demonstrating a 1\% average performance improvement. Table \ref{tab:exp-baselines} compares these results with other inference acceleration baselines, which, despite achieving some speedup, exhibit noticeable performance degradation.

\begin{table}[htp]
\centering
\small
\begin{tabular}{c|ccc}
\hline 
{Model} & {MMLU} & {GSM8K} & {Speedup} \cr
\hline
{Qwen2-7B } & {70.3} & {79.9} & {-} \cr
{L0-MoE } & {\textbf{70.4}} & {\textbf{80.5}} & {2.5x} \cr
\hline
{GPTQ} & {67.8} & {73.8} & {1.8x} \cr
{LLM Shearing} & {68.2} & {75.5} & {2.6x} \cr
{RKD + CoT} & {61.2} & {60.2} & {\textbf{5.1x}} \cr
\hline
\end{tabular}
\caption{Comparison with other inference acceleration baselines. We employ Qwen2-7B as the base LLM.}
\label{tab:exp-baselines}
\vspace{-3mm}

\end{table}
\begin{table}[htp]
\centering
\small
\begin{tabular}{ccc}
\hline 
{Model} & {MMLU} & {GSM8K} \cr
\hline
{L0-MoE} & {\textbf{70.4}} & {\textbf{80.5}}  \cr
\hline
{CCM w/o K-means} & {68.2} & {78.1}  \cr
{w/ random order batching} & {68.2} & {75.5} \cr
{w/ random batch batching} & {66.6} & {77.1}  \cr
\hline
{Random MoE} & {48.1} & {69.6} \cr
{Magnitude} & {52.6} & {69.1}  \cr
{OBS} & {68.4} & {74.1} \cr
{SVD} & {55.2} & {73.8}  \cr
\hline
\end{tabular}
\caption{Ablation study of CCM and L0-regularization. We conduct experiments on MMLU and GSM8K datasets with Qwen2-7B.}
\label{tab:exp-ablations}
\vspace{-5mm}
\end{table}

\vspace{-4pt}
\subsection{Ablation Study}

Table \ref{tab:exp-ablations} presents the ablation study on the CCM module, dynamic batching, and L0-regularization. Removing the K-means clustering from the CCM module results in a performance decline, underscoring the importance of effective sub-dataset curation. For dynamic batching, substituting it with random order or random batch scheduling also leads to degraded performance. 

In the context of MoE expert construction, we replace L0-regularization with four alternative methods: 1) \textbf{Random MoE} \cite{zhu-etal-2024-llama}: Selects MLP dimensions randomly, serving as a baseline to assess the necessity and effectiveness of dimension selection in expert construction, 2) \textbf{Magnitude} \cite{wandaprune-sun2023simple}): Selects the most influential elements in the weight matrix, improving upon traditional magnitude pruning by considering both the weights and their corresponding input activations using the L2 norm, 3) \textbf{OBS} \cite{obs-frantar2021m,obs-frantar2022optimal}): Identifies the most critical dimensions using the OBS Hessian matrix, which encapsulates second-order derivative information of the loss function with respect to model parameters. This approach is crucial for both pruning and quantization, as it helps retain the weights that most significantly impact model performance, 4) \textbf{SVD} \cite{svdllm-wang2024svd}): Decomposes the weight matrix using singular value decomposition and selects the most significant columns. By retaining the largest singular values, it reduces parameter count while preserving essential information. This truncation minimizes compression loss, and layer-wise updates further fine-tune the model to maintain accuracy. The results demonstrate the superiority of L0-MoE over them.

\begin{table}[htp]
\centering
\small
\begin{tabular}{c|ccc}
\hline 
{Model} & {MMLU}  & \#Para.(B) & {Speedup}  \cr
\hline
L0-MoE & {\textbf{70.4}} & {\textbf{23.3}}  & {2.5x} \cr
\hline
CCM Iter. ($Q=2$) & {68.2} & {23.3}  & {2.5x} \cr
CCM Iter. ($Q=5$) & {68.8} & {23.3}  & {2.5x} \cr
CCM Iter. ($Q=10$) & {69.7} & {23.3}  & {2.5x} \cr
\hline
Expert size ($K=8$) & {50.9} & {4.8}  & {\textbf{4.6x}}\cr
Expert size ($K=32$) & {69.2} & {12.7}  & {3.2x} \cr
\end{tabular}
\caption{Hyper-parameter tuning of sampling iterations ($Q$) and cluster size ($K$), keeping 2.8B activated parameters for L0-MoE. Qwen2-7B is the base LLM. \#Para.(B) is the number of model parameters.}
\label{tab:exp-kq-size}
\vspace{-5mm}
\end{table}

\vspace{-4pt}
\subsection{Discussion}

Our approach involves two critical hyperparameters: the sampling iterations $Q$ in CCM curated datasets and the expert size $K$ in MoE. Table \ref{tab:exp-kq-size} provides a detailed overview of hyperparameter tuning. Increasing the number of iterations for CCM enhances performance but also demands greater computational resources. We find that an initial iteration plus 20 additional iterations suffice to optimize model performance. While increasing the number of experts improves performance, it also reduces inference speed. Therefore, we select an appropriate expert size to balance performance enhancement and LLM acceleration. 

Besides, the optimal number of clusters ($K$) primarily depends on the characteristics of the pre-training corpus; thus, it may not directly transfer to other experiments if the corpus differs significantly. We recommend applying our method to pre-training corpora with abundant topical diversity, such as RedPajama, which contains millions of domains. In such corpora, a larger $K$ can effectively cluster more specialized subsets, enabling CCM to be more effectively applied when constructing L0-MoE models.

To assess model size impact, we conducted further experiments using a 1.5B-parameter Qwen2 model with 64 experts, achieving a 2x speedup without performance degradation. However, due to computational resource constraints, we have not yet experimented with larger models (e.g., 70B parameters). We hypothesize that larger LLMs could potentially achieve even greater speedups. We leave the verification of this hypothesis for larger-scale models as future work.

\vspace{-4pt}
\section{Conclusion and Future Work}
\vspace{-4pt}
In this paper, we propose a novel Mixture-of-Experts based approach to accelerate LLM inference, leveraging clustering confusion matrix for dataset curation, L0-regularization for expert selection, and dynamic batching for efficient training with only 30B tokens. Our method achieves a 2.5× speedup over dense LLMs, outperforming strong baselines nearly without performance loss. Future work will explore scaling our approach to larger LLMs and expanding the corpus size to further enhance L0-MoE performance beyond dense LLMs.

\section*{Limitations}


We did not compare our method with MoEs such as DeepSeek-MoE \cite{dai2024deepseekmoeultimateexpertspecialization}, Qwen-MoE \cite{qwen_moe}, and Mixtral \cite{jiang2024mixtralexperts}, which scale up model parameters in dense models with immense computational costs, processing trillions of tokens. In contrast, our approach utilizes only 30B tokens, making it more comparable to baseline post-training inference speedup methods.

Despite the promising results, several limitations remain: (1) The dataset for training each expert is selected via sequence-level semantic clustering, introducing exposure bias since MoE expert selection is performed at the token level. (2) The method does not explicitly measure inter-expert differences, potentially leading to redundant parameters that hinder L0-MoE’s inference acceleration. Future work should explore token-level dataset partitioning to mitigate exposure bias. Additionally, novel learning paradigms are needed to reduce parameter redundancy and enhance expert routing efficiency.

\bibliography{ref}

\clearpage

\appendix
\section{Appendix}
\label{sec:appendix}

This section provides further details on the model architecture, experimental setup (including evaluation tasks, baselines, and hyperparameter settings), and implementation details.

\subsection{Model Architecture}
\label{appendix:model-arch}
Table \ref{tab:model-arch} presents the detailed architecture of the baseline models and L0-MoE. All models incorporate group query attention (GQA) \cite{ainslie2023gqa} within the self-attention layer. For the L0-MoE models, the bottom 4 layers (for Qwen2) and 8 layers (for Mistral and Llama3) are configured as dense layers, while the remaining layers are transformed into MoE layers. We select the top-2 experts for each input token.

\subsection{Experimental Setups}
\label{appendix:experimental-setups}
\noindent \textbf{Evaluation Tasks.} We assess performance on four public benchmarks: MMLU \cite{hendrycks2021measuring}, GSM8K \cite{cobbe2021trainingverifierssolvemath}, HumanEval \cite{chen2021evaluatinglargelanguagemodels}, and BigBench Hard (BBH) \cite{suzgun-etal-2023-challenging}.

\begin{itemize}
 \item \textbf{MMLU} (Massive Multitask Language Understanding) \cite{hendrycks2021measuring} comprises 57 tasks spanning diverse subjects, including STEM (Science, Technology, Engineering, and Mathematics), humanities, social sciences, and specialized domains such as law and ethics.  
 \item \textbf{GSM8K} (Grade School Math 8K) \cite{cobbe2021trainingverifierssolvemath} is a benchmark designed to assess the mathematical reasoning capabilities of LLMs, containing 8,500 high-quality elementary math word problems.  
 \item \textbf{HumanEval} \cite{chen2021evaluatinglargelanguagemodels} evaluates the code generation capabilities of LLMs through 164 programming tasks, each requiring the model to generate a function that satisfies a given set of test cases.  
 \item \textbf{BBH} (BIG-Bench Hard) \cite{suzgun-etal-2023-challenging} is a subset of the larger BIG-Bench dataset, consisting of 23 highly challenging tasks designed to exceed the capabilities of current LLMs. These tasks demand creative problem-solving and deep domain expertise.  
\end{itemize}

\noindent \textbf{Baselines.} We compare the L0-regularized MoEs with the original LLMs and other LLM inference optimization methods, including the quantization baseline GPTQ \cite{frantar2022gptq}, the model pruning baseline LLM Shearing \cite{llmshearing-xia2023sheared}, and the knowledge distillation baseline RKD + CoT \cite{rkl-gu2024minillm, cotdistill-feng2024teaching}.  
\begin{itemize}  
\item \textbf{GPTQ} \cite{frantar2022gptq} is a block-wise quantization method that extends traditional power-of-two quantization by allowing non-uniform bin widths, enabling a better approximation of the original floating-point value distribution.  
\item \textbf{LLM-Shearing} \cite{llmshearing-xia2023sheared} employs structured pruning to construct lightweight, structured LLMs from pretrained checkpoints. It jointly removes attention heads, layers, feed-forward networks (FFNs), and hidden dimensions in an end-to-end manner to optimize efficiency.  
\item \textbf{RKD + CoT}: We apply RKD \cite{rkl-gu2024minillm} to distill the CoT \cite{cotdistill-feng2024teaching} capabilities of Qwen2-7B into Qwen2-1.5B. RKD \cite{rkl-gu2024minillm} aligns the student model with the teacher’s distribution using reverse Kullback-Leibler divergence (KLD), encouraging the student to focus on the most probable outcomes. This helps preserve the quality of the student model's predictions by distilling Chain-of-Thought (CoT) reasoning from the teacher model.  
\end{itemize}
 
\noindent \textbf{Hyper-parameter Setting.} The detailed hyper-parameter settings are presented in Table \ref{tab:exp-setups}. This includes the hyper-parameters for the clustering confusion matrix (CCM) as well as those for MoE training.

\subsection{Comparison with DuQuant}
To further validate our method, we compare it with DuQuant \cite{lin2024duquant}, a recent quantization technique targeting outlier activations in large language models (LLMs). DuQuant uses rotation and permutation to redistribute outliers, aiming to simplify quantization and improve robustness. We evaluate it on the LLama-3-8B model using MMLU \cite{hendrycks2021measuring} and GSM8K \cite{cobbe2021trainingverifierssolvemath} benchmarks. As shown in Table~\ref{tab:quant_results}, DuQuant suffers noticeable performance degradation, highlighting its limitations. In contrast, our L0-MoE method performs better under the same setting, demonstrating superior accuracy preservation.

\begin{table*} 
\setlength{\tabcolsep}{3pt}
\centering
\begin{tabular}{c|cccccccc}
\hline
{Model} &  {Parameters(B)} & {Layer} & {Hidden} & {Q/KV} & {FFN} & {MoE FFN} & {Experts}  \cr
 \hline

{Llama-3-8B} & {7.5} & {32} & {4096} & {32/8} & {14336}\cr
{Llama-3-8B w/ L0-MoE} & {25.1/3.2} & {32/24} & {4096} & {32/8} & {14336} & {1024} & {64:2}\cr
\hline
{Mistral-7B}  & {7.1} & {32} & {4096} & {32/8} & {14336} \cr
{Mistral-7B w/ L0-MoE}  & {24.7/2.9} & {32/24} & {4096} & {32/8} & {14336} & {1024} & {64:2} \cr
\hline
{Qwen2-7B}  & {7} & {28} & {3584} & {28/4} & {18944}\cr
{Qwen2-7B w/ L0-MoE}  & {23.3/2.8} & {28/24} & {3584} & {28/4} & {18944} & {1280} & {64:2} \cr

\hline
\end{tabular}
\caption{Detailed model architecture parameters. We denote the total and activated parameters of the MoEs, as well as the total layers and MoE layers, using the format ``32/24'', etc. All models utilize GQA, and we present the query/key-value heads. ``FFN'' refers to the dense decoder MLP size, while ``MoE FFN'' indicates the intermediate size of the expert for the MoE layer. The total and activated experts are represented as ``64:2'', etc.}
\label{tab:model-arch}
\end{table*} 

\begin{table*}
\centering
\begin{tabular}{lccc}
\toprule
\textbf{Model} & \textbf{MMLU} & \textbf{GSM8K} & \textbf{SpeedUp} \\
\midrule
Llama-3-8B & 66.6 & 56.0 & - \\
Llama-3-8B w/ L0-MoE & 66.3 & 55.9 & 2.0$\times$ \\
DeQuant/W4A4 & 57.9 & 51.6 & 0.46$\times$ \\
DeQuant + LWC/W4A4 & 62.2 & 51.1 & 0.46$\times$ \\
\bottomrule
\end{tabular}
\caption{Performance comparison of L0-MoE and DuQuant on MMLU and GSM8K benchmarks. Llama-3-8B is the base LLM. Due to the lack of DuQuant support in SGlang, we tested inference speed using naive PyTorch transformers (batch size = 64, sequence length = 1200). Without optimized kernels, DuQuant is slow (0.46x speedup), but future SGlang support could make it comparable to GPTQ (1.8x speedup).}
\label{tab:quant_results}
\end{table*}

\begin{table*}
\centering
\renewcommand{\arraystretch}{0.9}
\begin{tabular}{p{3.8cm}p{3.2cm}}
\hline 
\multicolumn{2}{c}{\textbf{CCM Hyper-parameters}} \\
\hline
$Q$ & 21 \\
$K$ & 64 \\
$d_{sv}$ & 1024 \\
$D_{sl}^{(0)}$ & \makecell[l]{12B tokens; \\ $|D_{s_i}^{(0)}| \approx 0.15B$} \\
$D_{sl}^{(l)}$, $l \ge 1$ & \makecell[l]{0.9B tokens; \\ $|D_{s_i}^{(0)}| \approx 0.007B$} \\
\hline
\multicolumn{2}{c}{\textbf{MoE Training Hyper-parameters}} \\
\hline
Sequence length & 4096 \\
Learning rate & {1e-4} \\
Warmup ratio (expert) & 0.2 \\
Warmup ratio (MoE) & 0.06 \\
Warmup type & Linear \\
Annealing ratio & 0.1 \\
Annealing type & Cosine \\
Batch tokens & 512K \\
$\alpha$ in Eq.~\ref{eq:train-moe} & 0.01 \\
$\beta$ in Equation \ref{eq:dsd} & {0.5} \\
$\lambda$ in Eq.~\ref{eq:aux-loss} & 0.3 \\
Training epoch & 1 \\
\hline
\end{tabular}
\caption{Hyper-parameters for CCM and MoE training.}
\label{tab:exp-setups}
\end{table*}

\end{document}